\documentclass[10pt,twocolumn,letterpaper]{article}

\usepackage{cvpr}              
\definecolor{cvprblue}{rgb}{0.21,0.49,0.74}
\usepackage[pagebackref,breaklinks,colorlinks,allcolors=cvprblue]{hyperref}

\newcommand{\hlg}[1]{{\color{green!50!black}{#1}}}

\usepackage{multirow}
\usepackage{bbding}
\usepackage{amsmath}
\usepackage[capitalize]{cleveref}
\usepackage{colortbl}
\usepackage{booktabs}
\usepackage{xcolor}
\usepackage{amssymb}
\usepackage{subcaption}
\usepackage{algorithm}
\usepackage{algorithmic}
\usepackage{newfloat}
\usepackage{listings}
\usepackage[accsupp]{axessibility}

\def\paperID{35040} 
\def\confName{CVPR}
\def\confYear{2026}

\title{TopoHR: Hierarchical Centerline Representation for Cyclic Topology Reasoning in Driving Scenes with Point-to-Instance Relations}

\author{Yifeng Bai$^{1,2}$\thanks{Equal contribution.} \quad Zhirong Chen$^{3}$\footnotemark[1] \quad Bo Song$^{1}$ \quad Erkang Cheng$^{3}$\thanks{Corresponding author. chengerkang@nullmax.ai.} \quad Haibin Ling$^{4}$ \\
$^{1}$Institute of Intelligent Machines, HFIPS, Chinese Academy of Sciences\\
$^{2}$University of Science and Technology of China \quad $^{3}$NullMax \quad $^{4}$Westlake University\\
}

\begin{document}

\maketitle
\begin{abstract}
Topology reasoning is crucial for autonomous driving. Current methods primarily focus on instance-level learning for centerline detection, followed by a sequential module for topology reasoning that relies on simplified MLP layers.
Moreover, they often neglect the importance of \textit{point-to-instance} (P2I) relationships in topology reasoning.
To address these limitations, we present TopoHR (Topological Hierarchical Representation), a novel end-to-end framework that establishes cyclic interaction between centerline detection and topology reasoning, allowing them to iteratively enhance each other. 
Specifically, we introduce a hierarchical centerline representation including point queries, instance queries, and semantic representations. These multi-level features are seamlessly integrated and fused within a hierarchical centerline decoder.
Furthermore, we design a hierarchical topology reasoning module that captures both fine-grained P2I relationships and global instance-to-instance (I2I) connections within a unified architecture. 
With these novel components, TopoHR ensures accurate and robust topology reasoning. 
On the OpenLane-V2 benchmark, TopoHR refreshes state-of-the-art performance with significant improvements. Notably, compared with previous best results, TopoHR achieves +3.8 in $\mathrm{DET}_{\text{l}}$, +5.4 in $\mathrm{TOP}_{\text{ll}}$ on $\text{subset\_A}$ and +11.0 in $\mathrm{DET}_{\text{l}}$, +7.9 in $\mathrm{TOP}_{\text{ll}}$ on $\text{subset\_B}$, validating the effectiveness of the proposed components.
The code will be shared publicly at \url{https://github.com/Yifeng-Bai/TopoHR.git}.
\end{abstract}
\section{Introduction}
\label{sec:intro}

\begin{figure}[t]
	\centering
	\includegraphics[width=0.975\linewidth]{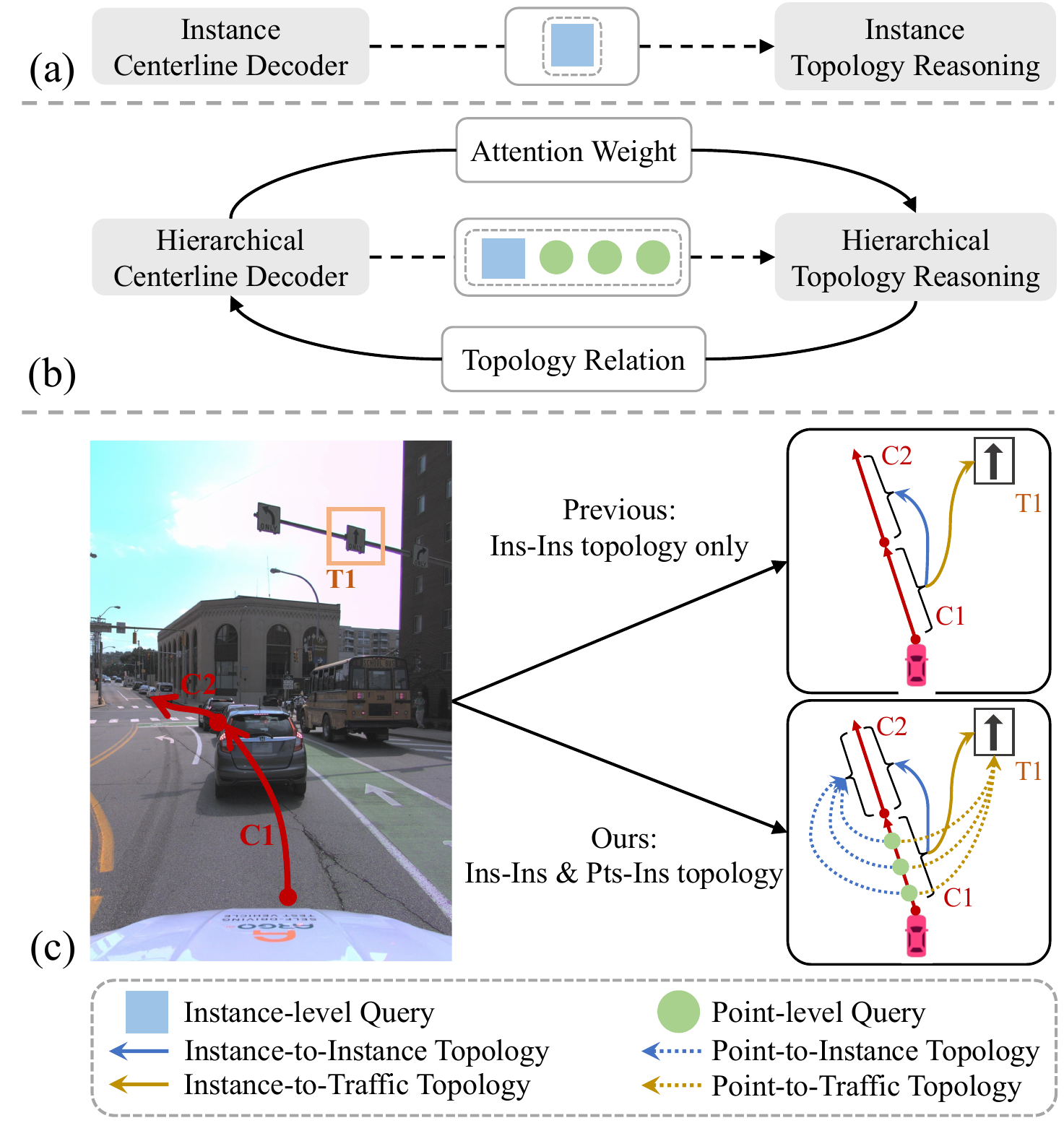}
	\caption{Different centerline detection and topology reasoning pipelines. (a) The conventional sequential pipeline. (b) Our cyclic pipeline. (c) Our motivation to integrate point-to-instance (P2I) and point-to-traffic (P2T) relations to boost topology reasoning.}  
	\label{fig:representation}
\end{figure}

Topological relationships of traffic scene elements refer to the fundamental connectivity rules governing road structures, hence are crucial for high-level autonomous driving functions.
Traditional lane detection methods~\cite{garnett20193d} and online mapping techniques~\cite{liaomaptr} focus on geometric accuracy but fail to capture scene topology relations. While HD Map can provide such topological information, it struggles with issues of freshness and scalability.
Recent methods~\cite{ligraph,wu2023topomlp,kalfaoglu2023topomask,kalfaoglu2025topobda,li2024lanesegnet} adopt a sequential pipeline (Fig.~\ref{fig:representation}(a)). In this pipeline, a centerline detector first extracts the instance-level centerline representation, and a topology reasoning module then analyzes the extracted centerline instances.

Previous centerline detection methods typically use instance-level query representations in a transformer decoder, defining the task as point-set prediction~\cite{ligraph,wu2023topomlp,fu2024topologic,lv2025t2sg,li2024lanesegnet}, curve parameter estimation~\cite{kalfaoglu2024topomaskv2,kalfaoglu2025topobda}, or binary segmentation~\cite{kalfaoglu2023topomask}. For instance, TopoNet~\cite{ligraph} represents centerlines as a point set and refines them using a Scene Graph Neural Network (SGNN). 
Alternatively, segmentation-based approaches like TopoMask~\cite{kalfaoglu2024topomaskv2} incorporate direction prediction to derive vectorized results from segmentation outputs. 
However, centerlines are inherently invisible, making it challenging to accurately extract their features through direct segmentation modeling. Current implementations either use segmentation as auxiliary supervision or rely on post-processing to generate vectorized point sets, which limits the full utilization of the rich information contained in segmentation results.

Building on instance-level queries from the centerline detector, a topology reasoning module usually applies MLP functions~\cite{ligraph, wu2023topomlp, lv2025t2sg, fu2024topologic} to infer topological relationships. For instance, TopoNet~\cite{ligraph} employs three MLP layers to reason about topological relationships using refined centerline queries. TopoMLP~\cite{wu2023topomlp} highlights the importance of detector performance in the cascade structure and enhances topology reasoning by integrating position embeddings into the MLP layer. Similarly, TopoLogic~\cite{fu2024topologic} improves reasoning performance by incorporating centerline geometry priors to link two centerlines through their end points.
While these methods achieve promising results in topology reasoning for driving scenes, there remains a significant gap in the accuracy of topological relationship predictions. These approaches often rely on cascaded architectures, where the detection and relation reasoning modules are optimized separately, leading to inconsistent feature representations. Additionally, the use of simplified prediction heads (e.g., MLPs) fails to adequately capture the complex spatial dependencies inherent in urban road networks.

\begin{figure*}[htb]
    \centering
    \includegraphics[width=0.875\linewidth]{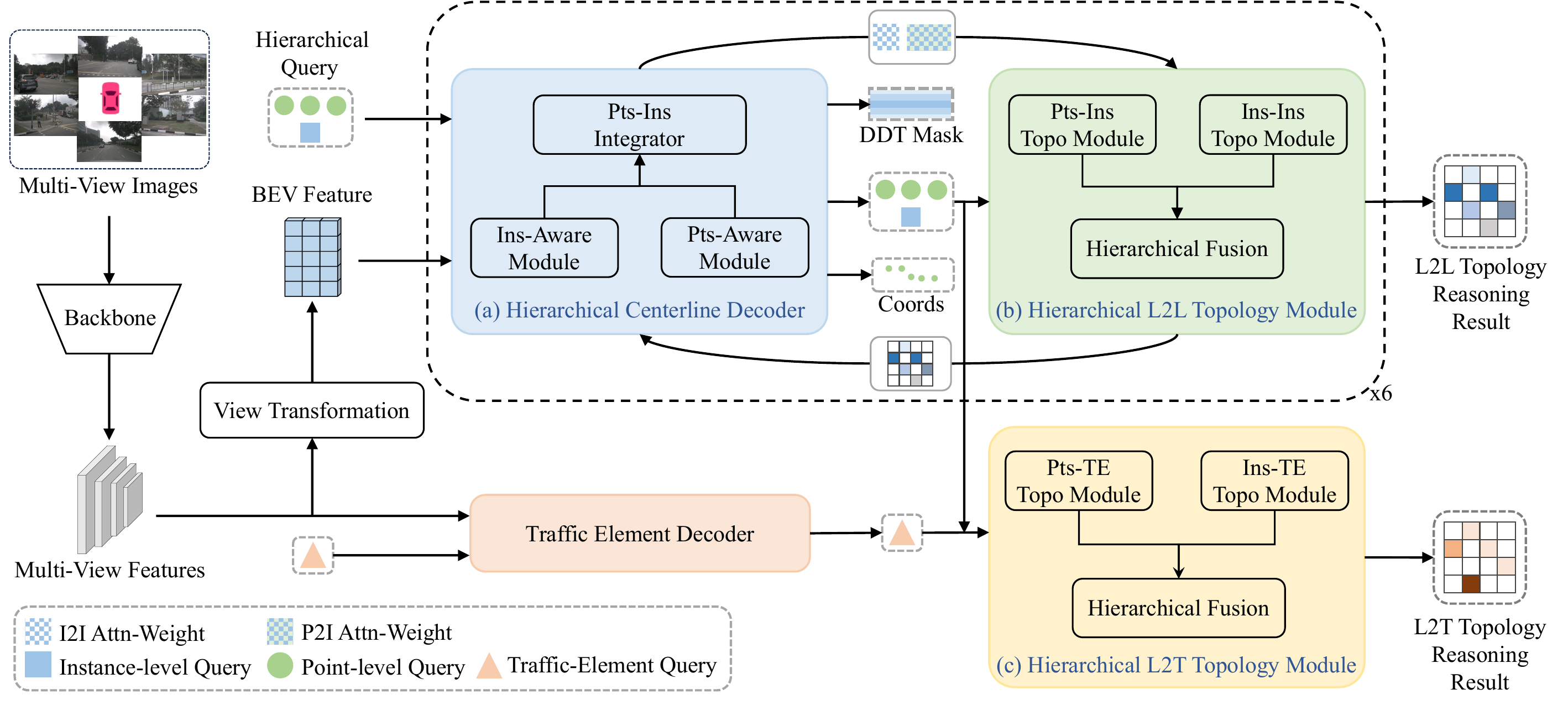}
    \caption{Overview of TopoHR. Aside from a BEV feature extractor and a traffic element decoder, TopoHR has three notable components: (a) Hierarchical Centerline Decoder, which employs a hierarchical query representation to simultaneously model point-level and instance-level features through a series of attention mechanisms and integration modules; (b) Hierarchical L2L Topology Module and (c) Hierarchical L2T Topology Module, both of which perform topology reasoning by combining both fine-grained point-to-instance relationships and global instance-to-instance topological connections.}
    \label{fig:overview}
\end{figure*}

To address the above issues, we propose TopoHR, a novel end-to-end topology reasoning framework based on a cyclic interaction structure and hierarchical centerline representation. Unlike previous cascade architectures that heavily rely on detector performance, we show that the detector and the topology reasoning module can mutually enhance each other.
As shown in Fig.~\ref{fig:representation}(b), we introduce a cyclic reasoning architecture where query attention weights from the detector are fed into the reasoning module as feedforward signals, while instance relations from the topology reasoning module are fed back to the detector. 

In the centerline detector, we integrate multiple centerline representations (point queries, instance queries, and semantic instances) and introduce specialized feature interaction modules.
First, to simultaneously capture local and global features, we design a hierarchical integrator that facilitates information exchange between point-aware and instance-aware queries. Second, we introduce a discrete distance transform segmentation method to extract spatial centerline information, which is then used for instance-aware query interaction through masked-attention.

For topology reasoning, we design a hierarchical topology reasoning module that captures both fine-grained \textit{point-to-instance} (P2I) and global \textit{instance-to-instance} (I2I) topological connections. These connections are fused to predict the final topological relations, as shown in Fig.~\ref{fig:representation}(c).

In the experiments, we evaluate TopoHR on the OpenLane-V2 dataset~\cite{wang2024openlane}. The results demonstrate significant improvements over existing methods under comparable settings. 
Specifically, TopoHR achieves scores of 34.6 $\mathrm{TOP}_{\text{ll}}$ and 35.6 $\mathrm{TOP}_{\text{lt}}$ on $\text{subset\_A}$, and 39.7 $\mathrm{TOP}_{\text{ll}}$ and 28.0 $\mathrm{TOP}_{\text{lt}}$ on $\text{subset\_B}$, thereby setting new records on OpenLane-V2.
These results correspond to absolute improvements of 5.4 points in $\mathrm{TOP}_{\text{ll}}$ and 3.4 points in $\mathrm{TOP}_{\text{lt}}$ on $\text{subset\_A}$, and 7.9 points in $\mathrm{TOP}_{\text{ll}}$ and 2.2 points in $\mathrm{TOP}_{\text{lt}}$ on $\text{subset\_B}$ over previous methods.

Our main contributions can be summarized as follows:
\begin{itemize}
\item We propose TopoHR, a novel end-to-end framework for centerline detection and topology reasoning, where the detection and reasoning modules iteratively interact with each other to cyclically enhance both performances.

\item We introduce a hierarchical centerline representation, where multi-level representations are seamlessly integrated and fused within a hierarchical centerline decoder. 

\item We propose a hierarchical topology reasoning module to capture both fine-grained P2I relationships and global I2I connections, leading to accurate topology results.

\item Our method demonstrates promising performance on the OpenLane-V2 dataset, surpassing previous state-of-the-art models by significant margins.
\end{itemize}

\section{Related Work}


\textbf{Map Element and Centerline Representation}. 
Centerline representations in topology reasoning typically employ map element formulation in online map understanding. These representations can be broadly categorized into rasterization-based and vectorization-based methods. 

Rasterization-based approaches formulate the extraction of map elements as a dense segmentation task. For instance, HDMapNet~\cite{li2022hdmapnet} pioneers the segmentation-based paradigm by fusing multi-view camera images and LiDAR point clouds within a Bird’s-Eye View (BEV) encoder-decoder architecture, thereby predicting instance-level map elements as semantic masks.
To reduce dependency on multi-modal data, MGMap~\cite{liu2024mgmap} proposes a camera-only framework with multi-granularity decoding, enabling hierarchical segmentation of map components.
Mask2Map~\cite{choi2024mask2map} enhances point-level feature extraction by incorporating deformable attention into instance segmentation.
More recently, MapVR~\cite{zhang2023online} applies differentiable rasterization to vectorized outputs and exploits distance variation on raster maps to provide accurate, geometry-aware supervision, without incurring additional computation during inference.

In contrast, vectorization-based methods represent map elements as sparse point sets. VectorMapNet~\cite{liu2023vectormapnet} introduces a two-stage pipeline combining polyline detection with geometric refinement. MapTR~\cite{liaomaptr} proposes a unified permutation-equivalent modeling strategy, enabling end-to-end vectorized map learning without explicit point ordering constraints. HIMap~\cite{zhou2024himap} unifies hierarchical queries for joint detection of lanes and traffic elements, showing robust generalization across diverse datasets. PivotNet~\cite{ding2023pivotnet} leverages sparse pivots for polyline parameterization, thereby enhancing the representation of geometric structures.

\begin{figure*}[t]
	\centering
	\includegraphics[width=0.925\linewidth]{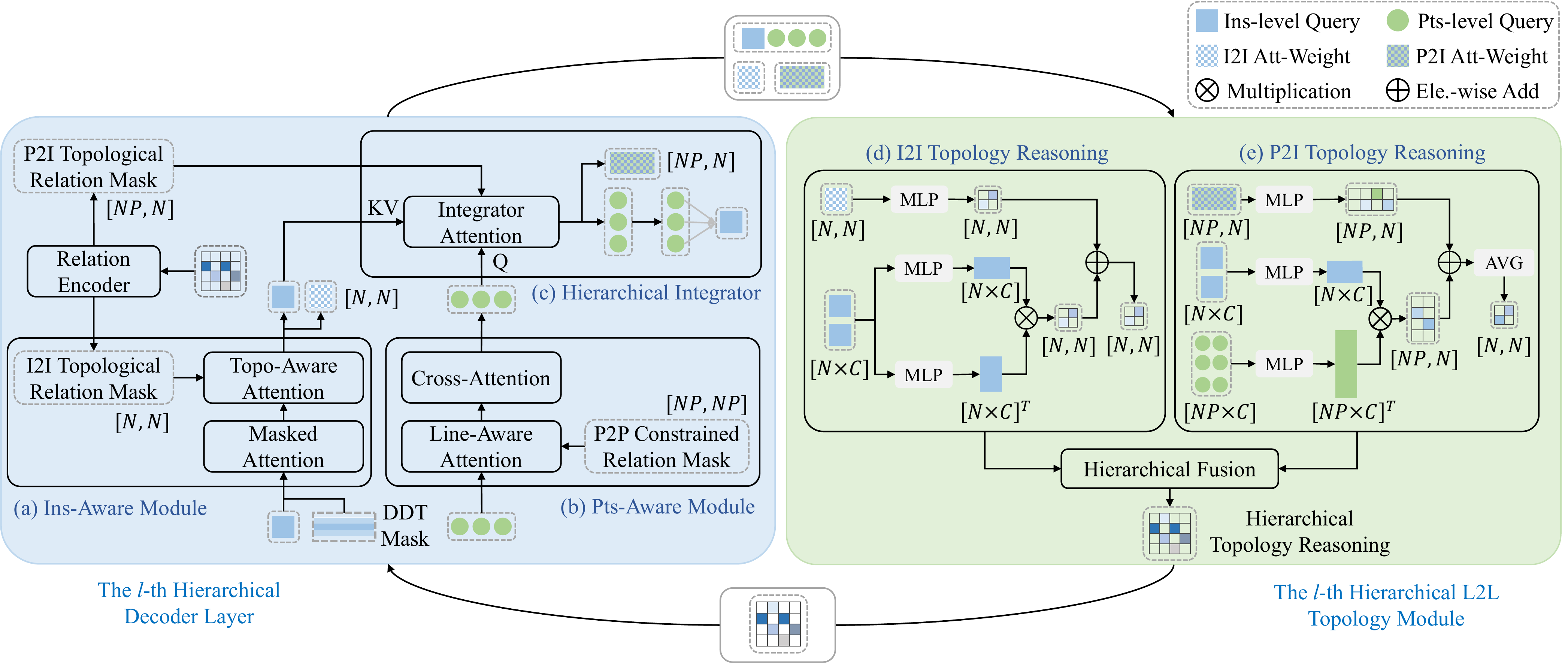}
	\caption{Illustration of the detailed architecture of hierarchical decoder Layer and hierarchical topology module: (a) an Instance-Aware Module, which combines a masked-attention module with a topo-aware attention to extract global semantic features; (b) a Point-Aware Module, which incorporates a line-aware attention module and a cross-attention module for capturing local geometric features; (c) a hierarchical Integrator, which integrates point-level and instance-level information through an integrator attention module and an aggregation module; (d) an Instance-to-Instance Topology Reasoning module; and (e) a Point-to-Instance Topology Reasoning module.}
	\label{fig:decoder layer}
\end{figure*}

\vspace{1mm}\noindent\textbf{Topology Detection and Reasoning}. 
Most topology reasoning approaches use a sequential pipeline, where a centerline detector is followed by a topology reasoning module.
For centerline detection, CenterLineDet~\cite{xu2023centerlinedet} represents centerlines as vertices and utilizes temporal feature fusion to achieve multi-camera perception. 
Beyond point-set inference, centerline detection also incorporates instance segmentation and curve parameterization. 
For instance, STSU~\cite{can2021structured} uses Bézier curves for centerlines. 
TopoMask~\cite{kalfaoglu2023topomask} introduces instance segmentation and adds direction prediction to parse vectorized results. 
TopoFormer~\cite{lv2025t2sg} utilizes the geometric distance between centerlines to guide global information aggregation and models plausible road structures under a counterfactual intervention layer.
In addition, SMERF~\cite{luo2024augmenting}, TopoSD~\cite{yang2024toposd} and SEPT~\cite{pei2025sept} propose the use of SD Map to enhance the centerline detector.

For topology reasoning, TopoNet~\cite{ligraph} first uses MLP to reduce instance embedding dimension, then sends the features to another MLP with sigmoid activation to predict their relationship.
TopoMLP~\cite{wu2023topomlp} follows the \textquotedblleft detection first, reasoning later\textquotedblright~rule to design a detector and an MLP that incorporates implicit position embedding.
TopoLogic~\cite{fu2024topologic} proposes a centerline geometry prior that explicitly preserves lane connectivity.
LaneSegNet~\cite{li2024lanesegnet} embeds topological affinity fields into instance segmentation, achieving real-time lane graph extraction.
Topo2Seq~\cite{yang2025topo2seq} converts the graph topology relation into a serialized representation and uses a hierarchical Transformer to achieve multiscale topological reasoning.
RelTopo~\cite{luo2025reltopo} constructs geometry-enhanced centerline relation by integrating positional embeddings and geometric distance embeddings.



\section{Method}

\subsection{Architecture Overview}
The proposed TopoHR framework is summarized in Fig.~\ref{fig:overview}. Starting from input multi-view images, it first extracts multi-view features through a shared feature backbone. These features are used to generate BEV features through view transformation and to detect traffic elements through a decoder. Then, TopoHR proceeds with three notable components: Hierarchical Centerline (HC) Decoder (Fig.~\ref{fig:overview}(a)), Hierarchical L2L Topology Module (Fig.~\ref{fig:overview}(b)) and Hierarchical L2T Topology Module (Fig.~\ref{fig:overview}(c)).
The HC Decoder and Hierarchical L2L Topology Module form a cyclic and iterative block, which takes BEV features as input and eventually produces L2L topology reasoning results. By contrast, the Hierarchical L2T Topology Module produces the L2T topology reasoning results. 

Specifically, the HC Decoder takes the hierarchical query representation $\mathbf{Q}_{\text{hcl}} \in \mathbb{R}^{N(P+1) \times C}$, where $N$, $P$, and $C$ denote the maximum number of centerlines, the number of points per centerline, and the number of channels, respectively. In contrast to previous approaches that rely solely on $N$ instance-level queries, TopoHR augments the representation with an additional set of $NP$ point-level queries, enabling joint modeling of global semantics local geometric details. The hierarchical query consists of two interrelated components: the point-level queries $\mathbf{Q}_{\text{pts}} \in \mathbb{R}^{P \times C}$ and the instance-level queries $\mathbf{Q}_{\text{ins}} \in \mathbb{R}^{C}$, which are refined through a series of attention mechanisms and a hierarchical integrator to enhance feature representations at both levels.

The hierarchical topology module performs topology reasoning by integrating updated hierarchical queries with P2I attention weights $\mathbf{W}_{\text{p2i}} \in \mathbb{R}^{NP \times N}$ from Integrator Attention and I2I attention weights $\mathbf{W}_{\text{i2i}} \in \mathbb{R}^{N \times N}$ from Topo-Aware Attention. This facilitates precise and semantically enriched topological structure understanding through both fine-grained point-level interactions and global instance-level relationships. The hierarchical L2T topology module adopts a similar design to extend the modeling to the relationships between centerlines and traffic elements.

\subsection{Iterative Cyclic Enhancement}
Conventional topology reasoning approaches optimize centerline detection and topology reasoning independently. Breaking this limitation, TopoLogic~\cite{fu2024topologic} incorporates centerline geometric priors with topology reasoning. 

Different from TopoLogic, TopoHR adopts a cyclic interaction mechanism that enables iterative refinement, shown in Fig.~\ref{fig:overview}. Specifically, P2I and I2I attention weights, $\mathbf{W}_{\text{p2i}}$ and $\mathbf{W}_{\text{i2i}}$, are passed from the hierarchical centerline decoder (Sec.~\ref{sec:hierarchical centerline decoder}) to the hierarchical topology reasoning module (Sec.~\ref{sec:hierarchical topology module}). Conversely, the topological relations $\mathbf{T}_{\text{p2i}}$ (P2I) and $\mathbf{T}_{\text{i2i}}$ (I2I) inferred by the topology reasoning module are fed back to the centerline decoder. This bidirectional information flow establishes a mutually reinforcing loop, consequently improving the overall performance.

\subsection{Hierarchical Centerline Decoder}
\label{sec:hierarchical centerline decoder}
The HC Decoder is designed to capture both local geometric details and global semantic information of centerline structures through a unified, multi-level representation, as illustrated in Fig.~\ref{fig:decoder layer}. Each centerline instance is modeled by a set of point-level queries encoding fine-grained spatial positions, as well as an instance-level query representing the overall semantic embedding of the entire centerline. By integrating information across these two levels, the decoder produces rich feature interactions via specialized attention mechanisms and facilitates effective aggregation and propagation of contextual cues. The architecture comprises three key modules, each tailored to address specific aspects of centerline detection and topology reasoning.

\begin{itemize}
    \item The \textbf{Instance-Aware Module} processes instance-level queries combining masked-attention and topo-aware attention. The masked-attention leverages a discrete distance transform mask (Section~\ref{sec:discrete distance transform}) to focus on relevant spatial regions, while the topology-aware attention utilizes the I2I topological relation mask to enhance global semantic feature extraction among centerline instances.

    \item The \textbf{Point-Aware Module} refines point-level queries by combining line-aware attention and cross-attention. The line-aware attention is constrained by a point-to-point (P2P) relation mask, enabling intra-instance interactions while preventing information leakage across different centerlines. The cross-attention integrates features from multiple points to enhance local geometric representation.

    \item The \textbf{Hierarchical Integrator Module} encodes both point- and instance-level information. It employs an integrator attention mechanism, where point-level queries interact with instance-level ones guided by the P2I topological relation mask. Then, an aggregation operation updates the instance-level queries based on the refined point-level features, ensuring coherent propagation of local and global information.
\end{itemize}

\subsubsection{Discrete Distance Transform Mask}
\label{sec:discrete distance transform}
Direct binary centerline segmentation presents inherent drawbacks: a centerline is artificially defined virtual object without distinct pixel-level features. Unlike segmentation labels, which are uniform for all positive samples, the distance transform encodes spatial proximity. It effectively addresses the challenge of centerline segmentation without relying on specific visual features.
Instead of directly using distance transform, we convert distance values to discrete ones to enhance computational efficiency and feature representation. Formally, given a centerline $\mathcal{C}=\left\{\mathbf{c}_1, \mathbf{c}_2, \ldots, \mathbf{c}_P\right\}$ of $P$ points, we first calculate the Euclidean distance from each pixel point $\mathbf{b}$ in the BEV map to the nearest centerline point $\mathbf{c}_i \in \mathcal{C}$. The distances are clipped within the half of the lane width $L_{\text{width}}$ and then normalized to the range $[0,1]$, and then discretized uniformly into 6 intervals. This produces the final discrete distance transform centerline mask $\texttt{DDT}(\mathbf{b})$. 

We apply the DDT mask to the masked-attention~\cite{cheng2022masked} in the instance-aware module and in instance matching and loss calculation. The DDT mask can better extract the position information of the centerline and converge more easily than the standard regression method.

\subsubsection{Hierarchical Relation Modeling}
\label{sec:hierarchical relation modeling}
\noindent\textbf{P2P Constrained Relation.} In the point-aware module (Figure~\ref{fig:decoder layer}(b)), we employ a line-aware attention mechanism to facilitate intra-instance interaction among point-level queries. Specifically, each instance-level query incorporates $P$ associated point-level queries (\textit{e.g.}, $P=11$), resulting in $NP$ point queries following sequential ordering. To construct the fixed P2P constrained relation $\mathbf{M_{\text{p2p}}} \in \mathbb{R}^{NP\times NP}$, we assign 0 to elements representing points within the same centerline instance and 1 to others. This matrix serves as an attention mask within the line-aware attention computation. Consequently, this mechanism constrains attention operations within individual centerline instances, blocks cross-instance interactions, and thus maintains instance-specific feature learning by preventing information exchange between different centerlines.

\vspace{1mm}\noindent\textbf{P2I \& I2I Topological Relation.} 
The inherent structural dependency and topological constraints in centerlines are particularly suitable for relation-aware feature learning. Thus motivated, we use relational learning to capture the relation between centerline detection and topology reasoning, as inspired by Relation DETR~\cite{hou2024relation}.

To use topology reasoning for improving the centerline detection, we introduce a relation encoder to process topological predictions from the previous layer. 
It employs 3-layer MLPs to generate two topological relations: (1) I2I topological  relation $\mathbf{M_{\text{i2i}}} \in \mathbb{R}^{N\times N}$ and (2) P2I topological relation $\mathbf{M_{\text{p2i}}} \in \mathbb{R}^{NP\times N}$. These relations are subsequently integrated as attention masks into the topo-aware attention and integrator-attention modules.
Our work represents the first attempt to establish a mutually reinforcing loop between centerline detection and topology reasoning, where each component iteratively enhances the performance of the other through information exchange and joint optimization.

\subsubsection{Hierarchical Integrator}

The hierarchical decoder facilitates multi-scale interactions between point-wise geometric features and instance-aware semantics in centerline modeling. Through a local-global attention architecture, it iteratively refines feature representations by fusing positional details with semantic contexts, enabling coherent feature propagation across scales.

As shown in the Figure~\ref{fig:decoder layer}(c), the hierarchical integrator module operates in two cross-level steps. Given point-level queries $\mathbf{Q}_{\text{pts}} \in \mathbb{R}^{NP \times C}$ updated from the point-aware module, and instance-level queries $\mathbf{Q}_{\text{ins}} \in \mathbb{R}^{N \times C}$ generated from the instance-aware module, the integrator runs in a cross-attention way. Specifically, point-level queries $\mathbf{Q}_{\text{pts}}$ act as $Q$, while instance-level queries $\mathbf{Q}_{\text{ins}}$ serve as $K$ and $V$. The integrator attention is formulated as:
\begin{equation}
    \begin{aligned}
        \mathbf{Q} = \mathbf{Q}_{\text{pts}} \mathbf{W}^Q, ~~ \mathbf{K} &= \mathbf{Q}_{\text{ins}}\mathbf{W}^K, ~~ \mathbf{V} = \mathbf{Q}_{\text{ins}} \mathbf{W}^V \\
        \mathbf{\hat Q}_{\text{pts}} &= \text{softmax}(\mathbf{Q\mathbf{K}}^T + \mathbf{M_{\text{p2i}}}) \mathbf{V},
    \end{aligned}
\label{eq:point-aware refinement}
\end{equation}
where $\mathbf{M_{\text{p2i}}}$ denotes the P2I topological relation. This allows each point-level query to selectively focus on relevant instance-aware features, thereby enriching its representation with both local and global contexts. The refined point-level queries are then aggregated into the updated instance-level queries through learnable coefficients $\mathbf{W_{\text{agg}}} \in \mathbb{R}^{P}$:
\begin{equation}
\begin{aligned}
\mathbf{\hat Q}_{\text{ins}} = \sum_{p=1}^P \mathbf{\hat Q}_{\text{pts}}{[:,p,:]} \ \text{softmax}(\mathbf{W_{\text{agg}}})_p.
\end{aligned}
\label{eq:instance_level aggregate}
\end{equation}
This design ensures that point-level queries are enriched with both local positional details and global semantic context, while instance-level queries are dynamically updated through aggregating refined point-aware features. 

\subsection{Hierarchical Topology Reasoning}
\label{sec:hierarchical topology module}
Existing topology reasoning approaches mostly depend on instance level query interactions, often modeled via simple MLPs. In contrast, we propose a hierarchical representation strategy that incorporates both point-to-instance and instance-to-instance topological relationships. Our key insight is grounded in the hierarchical nature of topological relationship: when considering the topology between two centerlines $\mathcal{C}_i$ and $\mathcal{C}_j$, the relationship must exist not only in the instance-level representation of the two centerlines, but also between the point-level representation of $\mathcal{C}_i$ and the instance-level representation of $\mathcal{C}_j$. This hierarchical perspective is equally applicable to the topological relationships between centerlines and traffic elements.

The hierarchical module comprises two prediction branches: (1) the I2I topology branch (Fig.~\ref{fig:decoder layer}(d)) that leverages instance-level queries and their self-attention weights through dual MLP encoding for similarity computation (inner product) and associative relationship extraction via attention-based MLP, and (2) the P2I topology branch (Fig.~\ref{fig:decoder layer}(e)) that jointly incorporates explicit feature correlations and latent dependencies. 
This design enables comprehensive topology reasoning by effectively integrating instance-aware features and their hidden dependencies. 

Specifically, the I2I prediction process is as follows:
\begin{equation}
    \begin{aligned}
        \mathbf{Q}^{\text{sim1}}_{\text{ins}}, \mathbf{Q}^{\text{sim2}}_{\text{ins}} &= \operatorname{MLP}(\mathbf{Q}_{\text{ins}}), \operatorname{MLP}(\mathbf{Q}_{\text{ins}}) \\
        \mathbf{T}_{\text{i2i}} &= (\mathbf{Q}^{\text{sim1}}_{\text{ins}}  (\mathbf{Q}^{\text{sim2}}_{\text{ins}})^\top) + \operatorname{MLP}(\mathbf{W}_{\text{i2i}}).
    \end{aligned}
\label{eq:ins2ins_topo_reasoning}
\end{equation}
The P2I prediction follows a similar workflow, except for an averaging operation along the point dimension:
\begin{equation}
    \begin{aligned}
        \mathbf{Q}^{\text{sim}}_{\text{pts}}, \mathbf{Q}^{\text{sim3}}_{\text{ins}} &= \operatorname{MLP}(\mathbf{Q}_{\text{pts}}), \operatorname{MLP}(\mathbf{Q}_{\text{ins}})\\
        \mathbf{T}_{\text{p2i}} &= (\mathbf{Q}^{\text{sim}}_{\text{pts}}  (\mathbf{Q}^{\text{sim3}}_{\text{ins}})^\top) + \operatorname{MLP}(\mathbf{W}_{\text{p2i}}).
    \end{aligned}
\label{eq:pts2ins_topo_reasoning}
\end{equation}
The final result is derived from the results of two hierarchical components $\mathbf{T}_{\text{i2i}}$ and $\mathbf{T}_{\text{p2i}}$.

\begin{table*}[t] 
  \begin{center}
    \fontsize{9pt}{9pt}\selectfont
    \renewcommand{\arraystretch}{1.05} 
    \caption{Performance comparison with other state-of-the-art methods on OpenLane-V2 subset\_A benchmark. All compared methods utilize ResNet-50 as the CNN backbone. The best results are highlighted in \textbf{bold}, the second best is \underline{underline}, and the third best is in \textit{italics}. (Repr: representation of centerline; \#Query: number of centerline queries.)}
    \begin{tabular}{c|c|c|c|ccccc}
    \toprule
    \textbf{Method}  & \textbf{Repr / \#Query} & \textbf{Epoch} & \textbf{SD Map} & $\mathrm{\textbf{DET}_{l}}$ & $\mathrm{\textbf{DET}_{\text{t}}}$ & $\mathrm{\textbf{TOP}_{ll}}$ & $\mathrm{\textbf{TOP}_{lt}}$ & $\mathrm{\textbf{OLS}}$ \\
      \midrule
      TopoNet~\cite{ligraph}  & Instance / 200 & 24 & - & \cellcolor[HTML]{DADADA}28.6 & 48.6 & \cellcolor[HTML]{DADADA}10.9 & \cellcolor[HTML]{DADADA}23.8 & \cellcolor[HTML]{DADADA}39.8\\
      SMERF~\cite{luo2024augmenting}  & Instance / 200 & 24 & \checkmark & \cellcolor[HTML]{DADADA}33.4 & 48.6 & \cellcolor[HTML]{DADADA}15.4 & \cellcolor[HTML]{DADADA}25.4 & \cellcolor[HTML]{DADADA}42.9 \\
      TopoLogic~\cite{fu2024topologic}  & Instance / 200 & 24 & - & \cellcolor[HTML]{DADADA}29.9 & 47.2 & \cellcolor[HTML]{DADADA}23.9 & \cellcolor[HTML]{DADADA}25.4 & \cellcolor[HTML]{DADADA}44.1 \\
      TopoLogic~\cite{fu2024topologic} & Instance / 200 & 24 & \checkmark & \cellcolor[HTML]{DADADA}34.4 & 48.3 & \cellcolor[HTML]{DADADA}28.9 & \cellcolor[HTML]{DADADA}28.7 & \cellcolor[HTML]{DADADA}47.5 \\
      TopoFormer~\cite{lv2025t2sg}  & Instance / 200 & 24 & - & \cellcolor[HTML]{DADADA}34.7 & 48.2 & \cellcolor[HTML]{DADADA}24.1 & \cellcolor[HTML]{DADADA}29.5 & \cellcolor[HTML]{DADADA}46.3 \\
      SEPT~\cite{pei2025sept} & Instance / 200 & 24 & \checkmark & \cellcolor[HTML]{DADADA}34.3 & \textit{48.9} & \cellcolor[HTML]{DADADA}31.2 & \cellcolor[HTML]{DADADA}29.7 & \cellcolor[HTML]{DADADA}48.4\\
      TopoPoint~\cite{futopopoint}  & Instance / 300 & 24 & - & \cellcolor[HTML]{DADADA}31.4 & \textbf{55.3} & \cellcolor[HTML]{DADADA}28.7 & \cellcolor[HTML]{DADADA}30.0 & \cellcolor[HTML]{DADADA}48.8 \\
      RelTopo~\cite{luo2025reltopo}  & Instance / 300 & 24 & - & \cellcolor[HTML]{DADADA}33.8 & \underline{50.9} & \cellcolor[HTML]{DADADA}29.2 & \cellcolor[HTML]{DADADA}32.2 & \cellcolor[HTML]{DADADA}48.9 \\
      \cmidrule{1-9} 
      TopoHR (Ours)  & Hierarchical / 200x(11+1) & 24 & - &  \cellcolor[HTML]{DADADA}\underline{36.1} & 48.3 & \cellcolor[HTML]{DADADA}\textit{31.8} & \cellcolor[HTML]{DADADA}\underline{34.6} &  \cellcolor[HTML]{DADADA}\underline{49.9} \\
      TopoHR-L (Ours)  & Hierarchical / 300x(11+1) & 24 & - & \cellcolor[HTML]{DADADA}\textit{35.6} & {48.8} & \cellcolor[HTML]{DADADA}\underline{32.3} & \cellcolor[HTML]{DADADA}\textit{34.3} & \cellcolor[HTML]{DADADA}\textit{49.8} \\
      TopoHR-L (Ours)  & Hierarchical / 300x(11+1) & 48 & - & \cellcolor[HTML]{DADADA}\textbf{37.6} & 47.0 & \cellcolor[HTML]{DADADA}\textbf{34.6} & \cellcolor[HTML]{DADADA}\textbf{35.6} & \cellcolor[HTML]{DADADA}\textbf{50.8} \\
      \bottomrule
    \end{tabular}
    \label{tab:results-openlanev2-subsetA}
  \end{center}
\end{table*}

\begin{table*}[t] 
  \begin{center}
    \fontsize{9pt}{9pt}\selectfont
    \renewcommand{\arraystretch}{1.05} 
    \caption{Performance comparison with other state-of-the-art methods on OpenLane-V2 subset\_B benchmark. All compared methods utilize ResNet-50 as the CNN backbone. The best results are highlighted in \textbf{bold}, the second best is \underline{underline}, and the third best is in \textit{italics}. (Repr: representation of centerline; \#Query: number of centerline queries.)}
    \begin{tabular}{c|c|c|ccccc}
    \toprule
      \textbf{Method}  & \textbf{Repr / \#Query} & \textbf{Epoch} & $\mathrm{\textbf{DET}_{l}}$ & $\mathrm{\textbf{DET}_{\text{t}}}$ & $\mathrm{\textbf{TOP}_{ll}}$ & $\mathrm{\textbf{TOP}_{lt}}$ & $\mathrm{\textbf{OLS}}$ \\
      \midrule
      TopoNet~\cite{ligraph}  & Instance / 200 & 24  & \cellcolor[HTML]{DADADA}24.3 & 55.0 & \cellcolor[HTML]{DADADA}6.7 & \cellcolor[HTML]{DADADA}16.7 & \cellcolor[HTML]{DADADA}36.8 \\
      TopoLogic~\cite{fu2024topologic}  & Instance / 200 & 24 & \cellcolor[HTML]{DADADA}25.9 & 54.7 & \cellcolor[HTML]{DADADA}21.6 & \cellcolor[HTML]{DADADA}17.9 & \cellcolor[HTML]{DADADA}42.3\\ 
      TopoFormer~\cite{lv2025t2sg}  & Instance / 200 & 24 & \cellcolor[HTML]{DADADA}34.8 & \underline{58.9} & \cellcolor[HTML]{DADADA}23.2 & \cellcolor[HTML]{DADADA}23.3 & \cellcolor[HTML]{DADADA}47.5 \\
      TopoPoint~\cite{futopopoint} & Instance / 300 & 24 & \cellcolor[HTML]{DADADA}31.2 & \textbf{60.2} & \cellcolor[HTML]{DADADA}28.3 & \cellcolor[HTML]{DADADA}\underline{27.1} & \cellcolor[HTML]{DADADA}49.2 \\
      RelTopo~\cite{luo2025reltopo} & Instance / 300 & 24 & \cellcolor[HTML]{DADADA}32.6 & \textit{58.8} & \cellcolor[HTML]{DADADA}31.8 & \cellcolor[HTML]{DADADA}25.8 & \cellcolor[HTML]{DADADA}\textit{49.7} \\
      \cmidrule{1-8} 
      TopoHR (Ours)  & Hierarchical / 200x(11+1) & 24 & \cellcolor[HTML]{DADADA}\textit{35.3} & 55.0 & \cellcolor[HTML]{DADADA}\textit{32.1} & \cellcolor[HTML]{DADADA}25.2 & \cellcolor[HTML]{DADADA}49.3\\
      TopoHR-L (Ours)  & Hierarchical / 300x(11+1) & 24 &  \cellcolor[HTML]{DADADA}\underline{38.4} & 53.5 & \cellcolor[HTML]{DADADA}\underline{35.8} & \cellcolor[HTML]{DADADA}\textit{26.6} & \cellcolor[HTML]{DADADA}\underline{50.8} \\
      TopoHR-L (Ours)  & Hierarchical / 300x(11+1) & 48 &  \cellcolor[HTML]{DADADA}\textbf{43.6} & 54.2 & \cellcolor[HTML]{DADADA}\textbf{39.7} & \cellcolor[HTML]{DADADA}\textbf{28.0} & \cellcolor[HTML]{DADADA}\textbf{53.4} \\
      \bottomrule
    \end{tabular}
    \label{tab:results-openlanev2-subsetB}
  \end{center}
\end{table*}

\subsection{Training Loss}
\noindent\textbf{Adaptive Topological Loss.} We propose an adaptive topological loss $\mathcal{L}_{\text{topo}}$ for topology reasoning. 
We employ a dynamic weighting strategy based on reparameterized cross-entropy, where negative sample weights follow exponential scaling $e^{\lambda_{\text{neg}}  x_i}$ with $x_i$ denoting predicted positive probability, while positive samples maintain fixed weighting $\lambda_{\text{pos}}$. This mechanism creates self-adaptive gradient modulation to put more penalties for negative samples with high confidence scores, effectively mitigating false positive predictions while preserving topological coherence.

In total, the training loss can be written as:
\begin{equation}
    \mathcal{L}=\mathcal{L}_{\text{det}}+\mathcal{L}_{\mathrm{\text{seg}}}+\mathcal{L}_{\mathrm{\text{topo}}},
\label{eq:total_loss}
\end{equation}
where $\mathcal{L}_{\text{det}}$ combines a focal loss~\cite{lin2018focal} for centerline detection and an $\ell_1$ loss for vectorized centerline regression; and $\mathcal{L}_{\text{seg}}$ combines a dice loss and a cross-entropy loss to guide the learning of instance-aware feature based on discrete distance transform centerline mask. 

\begin{figure}[t]
	\centering
	\includegraphics[width=0.95\linewidth]{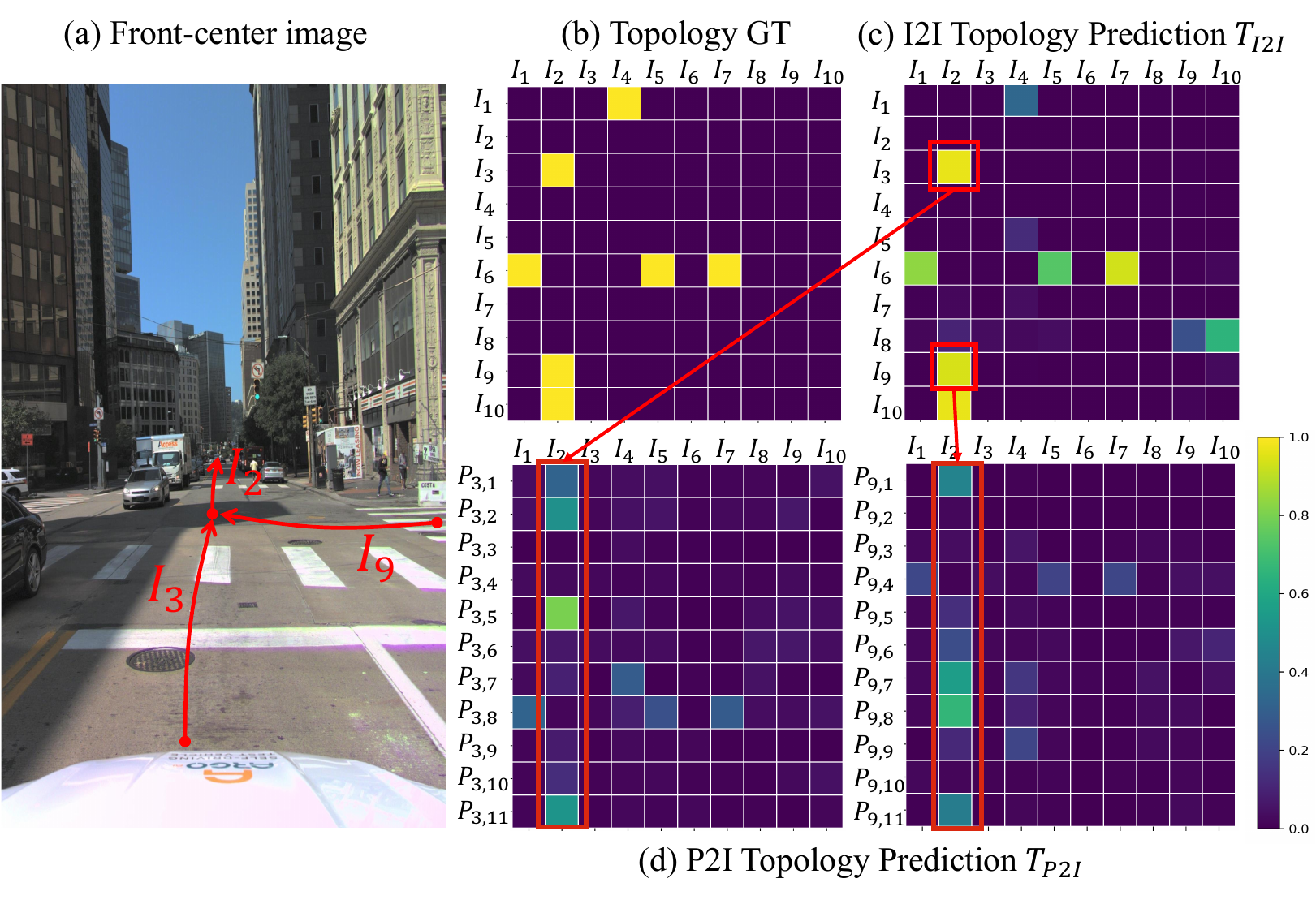}
	\caption{Instance-to-instance and point-to-instance topology reasoning results. (a) An input front-center image. (b) Groundtruth of centerline topology reasoning, where $(I_i, I_j) = 1$ denotes that the endpoint of the $I_i$-th centerline is connected to the start point of the $I_j$-th centerline. (c) Global instance-to-instance topology prediction. (d) Fine-grained point-to-instance topology prediction.}  
	\label{fig:heatmap}
\end{figure}    
\section{Experiments}

\subsection{Experimental Settings}

\noindent\textbf{Datasets and Metrics.}
%
TopoHR is evaluated on the OpenLane-V2 benchmark~\cite{wang2024openlane}, which comprehensively integrates Argoverse2~\cite{wilson2023argoverse} and nuScenes~\cite{caesar2020nuscenes}. It comprises 2,000 scenes, partitioned into $\text{subset\_A}$ and $\text{subset\_B}$, and provides multi-view imagery at 2 Hz along with detailed annotations for 3D centerlines, traffic elements, and their topological relationships. Specifically, $\text{subset\_A}$ consists of seven camera views, while $\text{subset\_B}$ includes six.

We adopt the official evaluation metrics as in~\cite{fu2024topologic}, including $\mathrm{DET}_{\text{l}}$ (mean Fréchet distance across matching thresholds), $\mathrm{DET}_{\text{t}}$ (IoU-based similarity for traffic elements), $\mathrm{TOP}_{\text{ll}}$ (similarity of the centerline topology matrix), and $\mathrm{TOP}_{\text{lt}}$ (centerline–traffic element topology similarity). The overall score ($\mathrm{OLS}$) is computed as their average. 

\vspace{.5mm}\noindent\textbf{Implementation.} 
We employ ResNet50~\cite{he2016deep} and Feature Pyramid Network (FPN)~\cite{lin2017feature} for multi-scale features. All input images are resized to $1024 \times 775$, as in most previous methods. Following BEVFormer~\cite{li2024bevformer}, we project image features into the pre-defined BEV space with grid resolution $200\times 100$. The DDT mask follows Mask2Former, initialized to zeros before the first decoder layer to avoid restricting interactions between instance queries and image features. Initial topology relations follow the TopoLogic geometric distance strategy using endpoint–startpoint distance mappings.
We initialize 200 hierarchical centerline queries for both centerline detection and topology reasoning. In the centerline detector, the regression head consists of 3-layer MLPs with LayerNorm and ReLU, outputs $11 \times 3$ 3D offsets per centerline. 
The proposed topology reasoning method focuses primarily on modeling the topological relationships of the centerlines. 
We follow the settings in~\cite{fu2024topologic} to detect traffic elements with no further changes.
For a fair comparison, TopoHR is trained using 8 NVIDIA 4090 GPUs with a total batch size of 8 over 24 epochs. For TopoHR-L, which requires higher GPU memory, experiments are conducted on 8 NVIDIA A100 GPUs. For optimization, we adapt the AdamW optimizer~\cite{loshchilov2017decoupled} with an initial learning rate of $3\times 10^{-4}$ and a weight decay of 0.01. It is worth noting that TopoHR achieves 12.6 FPS on a single RTX 4090.

\subsection{Comparisons with State-of-the-art Methods}
We evaluate TopoHR against state-of-arts on OpenLane-V2 (Table~\ref{tab:results-openlanev2-subsetA}). Both TopoHR and TopoHR-L substantially outperform existing methods on $\text{subset\_A}$ in $\mathrm{DET}_{\text{l}}$, $\mathrm{TOP}_{\text{ll}}$ and $\mathrm{TOP}_{\text{lt}}$. Notably, the TopoHR with 200 centerline queries achieves a $\mathrm{TOP}_{\text{ll}}$ score of 31.8, surpassing RelTopo~\cite{luo2025reltopo} by 2.6 points. 
Given the increased complexity of subset\_A, TopoHR-L exhibits limited performance gain during early training stages. To ensure full convergence of TopoHR-L, we extend its training to 48 epochs. As a result, TopoHR-L achieves improvements of 3.8 in $\mathrm{DET}_{\text{l}}$ and 5.4 in $\mathrm{TOP}_{\text{ll}}$. 
Since TopoHR employs the same traffic decoder as TopoLogic~\cite{fu2024topologic}, their performance in $\mathrm{DET}_{\text{t}}$ is nearly identical.
These results clearly demonstrate that our cyclic framework, combined with P2I relation modeling, significantly enhances both centerline detection and topology reasoning.

We further assess TopoHR on $\text{subset\_B}$ (Table~\ref{tab:results-openlanev2-subsetB}). TopoHR achieves outstanding performance in topology reasoning, attaining a notable 35.3 $\mathrm{DET}_{\text{l}}$ and outperforming RelTopo that utilizes a larger number of centerline queries.
Under the same number of centerline queries, TopoHR-L maintains a clear advantage over RelTopo, achieving 43.6 $\mathrm{DET}_{\text{l}}$ and 39.7 $\mathrm{TOP}_{\text{ll}}$. Additional experimental and qualitative results are provided in Suppl.\ Materials.

\begin{table}
    \centering
    \fontsize{9pt}{9pt}\selectfont
    \setlength{\tabcolsep}{2pt} 
    \renewcommand{\arraystretch}{1.0} 
    \caption{Ablation of hierarchical centerline representation. (P2P Constrain: P2P constrained relation mask; DT: distance transform mask; DDT: discrete distance transform mask.)}
    \begin{tabular}{c|ccc|cccc}
    \toprule
         \multirow{2}{*}{Repr} & \multicolumn{1}{c}{P2P} & \multicolumn{1}{c}{Hierarchical} &   Seg &   \multirow{2}{*}{$\mathrm{DET}_{\text{l}}$} &  \multirow{2}{*}{$\mathrm{TOP}_{\text{ll}}$}  \\
         & Constrain & Integrator & GT & & \\
         \midrule
         Ins & - & - & - & 26.8 & 23.1\\
         \cmidrule{1-6}
         Ins+Pts  & - & - & - & 19.6 & 24.9\\
         Ins+Pts  & \checkmark & - & - & 20.1 & 25.1\\
         Ins+Pts  & \checkmark & \checkmark & - & 32.2 & 26.3\\
         \cmidrule{1-6}
         Ins+Pts+Seg  & \checkmark & \checkmark & 0/1  &32.6 &29.8\\
         Ins+Pts+Seg  & \checkmark & \checkmark & DT &32.0 & 29.8\\
         \cellcolor[HTML]{DADADA}Ins+Pts+Seg & \cellcolor[HTML]{DADADA}\checkmark & \cellcolor[HTML]{DADADA}\checkmark & \cellcolor[HTML]{DADADA}DDT & \cellcolor[HTML]{DADADA}34.6 & \cellcolor[HTML]{DADADA}30.6\\
         \addlinespace
         \cmidrule{1-6}
         \hlg{\textit{Inprovement}} & - & - & - & \hlg{7.8$\uparrow$} & \hlg{7.5$\uparrow$}\\ 
    \bottomrule
    \end{tabular}
    \label{tab:ablation of Hierarchical Centerline Representation}
\end{table}

\subsection{Ablation Study}
\label{sec:ablation study}
Four ablation experiments are conducted on OpenLane-V2 $\text{subset\_A}$ to evaluate contributions of individual components. 
To construct a memory-efficient baseline, we adapt TopoLogic~\cite{fu2024topologic} by removing its GNN module. As TopoHR adopts the same traffic decoder as TopoLogic, traffic detection metrics are omitted from the ablation study.
All experiments use 200 centerline queries with focal loss for topology reasoning and are trained for 24 epochs by default.

\vspace{.5mm}\noindent\textbf{Hierarchical Centerline Representation.} 
We evaluate various centerline representations using the topology reasoning module of TopoLogic~\cite{fu2024topologic}. 
As reported in Table~\ref{tab:ablation of Hierarchical Centerline Representation}, incorporating point-level queries reduces $\mathrm{DET}_{\text{l}}$ to 19.6, while $\mathrm{TOP}_{\text{ll}}$ slightly increases. Introducing the point-to-point (P2P) constrained relation, which enables intra-instance interaction, leads to a modest performance improvement. Including the hierarchical integrator yields a significant boost, achieving 32.2 $\mathrm{DET}_{\text{l}}$ and 26.3 $\mathrm{TOP}_{\text{ll}}$, underscoring its critical role in ensuring coherent propagation of both local and global information. Further, we enhance the representation by incorporating masked-attention and supervision with a segmentation loss. Using binary segmentation GT achieves 32.6 $\mathrm{DET}_{\text{l}}$ and 29.8 $\mathrm{TOP}_{\text{ll}}$, while DT yields similar results. Our DDT achieves the best performance, bringing 7.8 $\mathrm{DET}_{\text{l}}$ and 7.5 $\mathrm{TOP}_{\text{ll}}$ gain over the baseline. These results collectively validate that each hierarchical component contributes meaningfully to the overall performance. It is also noteworthy that the incorporation of point-level queries and DDT segmentation within the hierarchical centerline representation only result in a 13.8\% increase in parameters.

\begin{table}
    \centering
    \fontsize{9pt}{9pt}\selectfont
    \setlength{\tabcolsep}{2.5pt} 
    \renewcommand{\arraystretch}{1.0} 
    \caption{Ablation of cyclic pipeline and P2I relation using hierarchical centerline representation and adaptive topological loss.}
    \begin{tabular}{cc|c|ccc}
    \toprule
    \multicolumn{2}{c|}{Forward} & \multicolumn{1}{c|}{Backward} & \multirow{2}{*}{$\mathrm{DET}_{\text{l}}$} & \multirow{2}{*}{$\mathrm{TOP}_{\text{ll}}$} & \multirow{2}{*}{$\mathrm{TOP}_{\text{lt}}$} \\ 
    Query & Weight & Topo &  & &  \\
    \midrule
    $\mathbf{Q}_{\text{ins}}$ & - & - & 34.8  & 31.0 & 31.1 \\
    \cmidrule{1-6}
    $\mathbf{Q}_{\text{ins}}$ & $\mathbf{W}_{\text{i2i}}$ & $\mathbf{T}_{\text{i2i}}$ & 35.6  & 31.5 & 32.9  \\
    \cmidrule{1-6}
    \cellcolor[HTML]{DADADA}$\mathbf{Q}_{\text{ins}}$+$\mathbf{Q}_{\text{pts}}$ & \cellcolor[HTML]{DADADA}$\mathbf{W}_{\text{i2i}}$+$\mathbf{W}_{\text{p2i}}$ & \cellcolor[HTML]{DADADA}$\mathbf{T}_{\text{i2i}}$+$\mathbf{T}_{\text{p2i}}$ & \cellcolor[HTML]{DADADA}36.1 & \cellcolor[HTML]{DADADA}31.8 & \cellcolor[HTML]{DADADA}34.6\\
    \cmidrule{1-6}
    \hlg{\textit{Inprovement}} & - & - & \hlg{1.3$\uparrow$} & \hlg{0.8$\uparrow$} & \hlg{3.5$\uparrow$}\\ 
    \bottomrule
    \end{tabular}
    \label{tab:ablation of cyclic pipeline}
\end{table}

\begin{table}
    \centering
    \fontsize{9pt}{9pt}\selectfont
    \setlength{\tabcolsep}{4pt} 
    \renewcommand{\arraystretch}{1.0} 
    \caption{Ablation of adaptive topological loss in TopoHR. (FC: focal loss; Dice: dice loss; ATL: adaptive topological loss.)}
    \begin{tabular}{c|cc|ccc}
    \toprule
        Topo Loss & $\lambda_{\text{neg}}$ & $\lambda_{\text{pos}}$ & $\mathrm{DET}_{\text{l}}$ & $\mathrm{TOP}_{\text{ll}}$  &   $\mathrm{TOP}_{\text{lt}}$\\
         \midrule
        FC& - & - & 34.9 & 31.2 & 32.3 \\ 
        Dice& - & - & 30.4 & 26.0 & 29.3  \\ 
        \cmidrule{1-6}
        ATL&  5   &  200  & 33.8   & 31.0  & 33.6  \\
        \cellcolor[HTML]{DADADA}ATL& \cellcolor[HTML]{DADADA}5 & \cellcolor[HTML]{DADADA}400 & \cellcolor[HTML]{DADADA}36.1  & \cellcolor[HTML]{DADADA}31.8  & \cellcolor[HTML]{DADADA}34.6  \\
        ATL&  5  &  800  & 35.7  & 31.3  & 33.9 \\
        ATL&  10  &  400  & 34.0  & 31.1  & 33.2  \\
        \cmidrule{1-6}
        \hlg{\textit{Inprovement}} & - & - & \hlg{1.2$\uparrow$} & \hlg{0.6$\uparrow$} & \hlg{2.3$\uparrow$}\\ 
    \bottomrule
    \end{tabular}
    \label{tab:ablation of Adaptive Topological Loss}
\end{table}

\vspace{.5mm}\noindent\textbf{Cyclic Pipeline and P2I Relation.} 
Table~\ref{tab:ablation of cyclic pipeline} demonstrates the effectiveness of cyclic information flow and hierarchical relation modeling, using our proposed hierarchical centerline representation and adaptive topological loss. Using only instance-level queries without cyclic feedback yields 34.8 $\mathrm{DET}_{\text{l}}$ and 31.0 $\mathrm{TOP}_{\text{ll}}$. Introducing I2I attention weights in the forward path and backward feedback of I2I topological result improves the $\mathrm{DET}_{\text{l}}$ and $\mathrm{TOP}_{\text{ll}}$ by 0.8 and 0.5, demonstrating the benefit of iterative cyclic refinement. Finally, incorporating both instance- and point-level queries, along with both I2I and P2I relations in the cyclic pipeline, achieves the best results of 36.1 $\mathrm{DET}_{\text{l}}$, 31.8 $\mathrm{TOP}_{\text{ll}}$ and 34.6 $\mathrm{TOP}_{\text{lt}}$. These results confirm that the proposed cyclic architecture, especially with P2I relation modeling, significantly enhances both detection and topology reasoning performance. We present detailed qualitative results in Figure~\ref{fig:heatmap}, which illustrate that point-to-instance topology prediction enables more fine-grained connectivity. These findings validate the hierarchical topology concept proposed in this work: topological relationships are manifested not only at the instance-level between centerlines, but also through their point-level representations.

\vspace{.5mm}\noindent\textbf{Adaptive Topological Loss (ATL).} 
The efficacy ATL  of our TopoHR is validated in Table~\ref{tab:ablation of Adaptive Topological Loss}. Replacing the standard Focal Loss with our proposed ATL results in improvements of 1.2 $\mathrm{DET}_{\text{l}}$ 0.6 $\mathrm{TOP}_{\text{ll}}$ and 2.3 $\mathrm{TOP}_{\text{lt}}$. 
We limit the exploration of $\lambda_{\text{neg}}$ and $\lambda_{\text{pos}}$ combinations, as our primary objective is to validate the importance of emphasizing false negative predictions in topological loss formulation rather than pursuing hyperparameter optimization.

\section{Conclusion}
We present TopoHR, an end-to-end topology reasoning approach integrating cyclic detector-topology interactions and hierarchical centerline representations. 
TopoHR overcomes the limitations of sequential pipelines by enabling detector-topology co-evolution, while the hierarchical representation fuses multi-level centerline features through discrete distance transform.  
A unified hierarchical topology module simultaneously captures fine-grained point-to-instance relationships and global topological connections. 
Extensive experiments on OpenLane-V2 clearly validates the effectiveness of design and components of TopoHR.

{
    \small
    \bibliographystyle{ieeenat_fullname}
    \bibliography{main}
}


\end{document}